\documentclass[conference]{IEEEtran}
\IEEEoverridecommandlockouts
\usepackage{amsmath,amssymb,amsfonts}
\usepackage{algorithmic}
\usepackage{graphicx}
\usepackage{textcomp}
\usepackage{xcolor}
\usepackage{dirtytalk}
\usepackage{hyperref}
\usepackage[
backend=biber,
style=ieee,
]{biblatex}
\begin{document}

\title{Modelling and Simulation of Neuromorphic Datasets for Anomaly Detection in Computer Vision
\thanks{This research is part of the Edgy Organism project, supported by an Engineering and Physical Sciences Research Council (EPSRC) grant (EP/Y030133/1).}
}

\author{
\textbf{Mike Middleton}\textsuperscript{1}, Teymoor Ali\textsuperscript{2}, Hakan Kayan\textsuperscript{3}, Basabdatta Sen Bhattacharya\textsuperscript{4},\\
Charith Perera\textsuperscript{3}, Oliver Rhodes\textsuperscript{4}, Elena Gheorghiu\textsuperscript{2}, Mark Vousden\textsuperscript{5}, Martin A. Trefzer\textsuperscript{1} \\\\
\textsuperscript{1}Bio-Inspired Systems and Technologies Lab, University of York, York, UK \\
\textsuperscript{2}Faculty of Natural Sciences, University of Stirling, Stirling, UK \\
\textsuperscript{3}School of Computer Science and Informatics, Cardiff University, Cardiff, UK \\
\textsuperscript{4}School of Electrical and Electronic Engineering, University of Manchester, Manchester, UK \\
\textsuperscript{5}Electronics and Computer Science, University of Southampton, Southampton, UK \\\\
\texttt{michael.middleton@york.ac.uk, t.r.ali@stir.ac.uk, kayanh@cardiff.ac.uk,}\\
\texttt{basab@manchester.ac.uk, m.vousden@soton.ac.uk,}\\
\texttt{pererac@cardiff.ac.uk, oliver.rhodes@manchester.ac.uk,}\\
\texttt{elena.gheorghiu@stir.ac.uk, martin.trefzer@york.ac.uk}
}

\maketitle

\begin{abstract}
    Limitations on the availability of Dynamic Vision Sensors (DVS) present a fundamental challenge to researchers of neuromorphic computer vision applications.  In response, datasets have been created by the research community, but often contain a limited number of samples or scenarios.  To address the lack of a comprehensive simulator of neuromorphic vision datasets, we introduce the Anomalous Neuromorphic Tool for Shapes (ANTShapes), a novel dataset simulation framework.  Built in the Unity engine, ANTShapes simulates abstract, configurable 3D scenes populated by objects displaying randomly-generated behaviours describing attributes such as motion and rotation.  The sampling of object behaviours, and the labelling of anomalously-acting objects, is a statistical process following central limit theorem principles.  Datasets containing an arbitrary number of samples can be created and exported from ANTShapes, along with accompanying label and frame data, through the adjustment of a limited number of parameters within the software.  ANTShapes addresses the limitations of data availability to researchers of event-based computer vision by allowing for the simulation of bespoke datasets to suit purposes including object recognition and localisation alongside anomaly detection.

\end{abstract}

\begin{IEEEkeywords}
Neuromorphic computing, Computer vision, Surveillance, Anomaly detection, Edge device, Dataset simulation
\end{IEEEkeywords}

\section{Background}


Anomaly detection using Artificial Neural Networks (ANNs) has attracted considerable research attention in recent years given the growing accessibility of deploying networks on devices such as GPUs \cite{ann1, ann2, ann3, ann4, ann5}.  However, given constraints on size, weight and power consumption, the applicability of such methods at the edge are limited.  To this end, neuromorphic Dynamic Vision Sensors (DVSs) have grown in popularity due to their remarkably low power consumption and their high dynamic ranges and rates of capture \cite{dvs1, dvs2, dvs_monochrome}.

Despite these improvements over frame-based cameras, there is a lack of reliable neuromorphic vision datasets captured by event-based cameras due to their relative scarcity \cite{esim}.  Specifically, neuromorphic vision datasets for anomaly detection are rare because neuromorphic sensors are not typically used as security cameras.  Neuromorphic vision datasets are desirable for the training of Spiking Neural Network (SNN) models for anomaly detection.  SNNs show promising results for autonomous anomaly identification with continuous learning at the edge and have attracted considerable research attention accordingly \cite{ucf_crime_dvs, snn1, snn2, snn3, snn4}. 

To address the lack of neuromorphic datasets, simulators have been developed to extrapolate events from frame data.  These methods interpolate between frames and simulate the pixel intensity detection of a DVS camera in software \cite{esim, realistic_dvs_sim}.  However, if the temporal capture rate of the source footage is too low to reliably encode the actions of agents in the scene, simulated event data will be unreliable.  Another approach is to point a DVS camera at a monitor displaying frame data \cite{ucf_crime_dvs}.  This method is imperfect and can introduce external noise into the capture if the DVS camera is disturbed during data conversion.

DVS camera behaviour has simulated to capture a virtual scene in CARLA, an application which simulates 3D urban environments from the point of view of a car \cite{carla}.  Its purpose is for the testing and development of autonomous vehicles rather than anomaly detection.  It is therefore left for the user to define the anomalies within generated datasets themselves.  This is open to human error and could take many hours of labour without a robust method of automation.  Furthermore, it is primarily designed for ego-motion, or motion of the self.  For anomaly detection in a stand-off scenario, such as information captured by a surveillance camera, a simulator where the camera remains fixed is needed.

These unresolved issues motivate the development of ANTShapes.  We aim to address the lack of parameterisable neuromorphic datasets by developing a comprehensive simulator of anomalous behaviour exhibited by objects in an abstract scene, approximating the view of a CCTV camera.  Rather than simulating a realistic environment as in CARLA, we aim to simulate abstract scenes for lower-level analysis of learning characteristics displayed by SNNs.  Neuromorphic learning presents challenges compared to traditional learning in ANNs \cite{snn_challenges}, so we aim to understand the potential of SNN anomaly detection in highly-controllable environments before progressing to realistic scenes.

\section{Definition of Anomalies}
\label{ss:anomalies} 

In ANTShapes, agents are objects in a scene that share similar properties and behaviours, such as shape, scale, texture and patterns of movement.  In a virtual model of a real-life scene, such as the view of a street captured by a vision sensor, agents could be categorised as pedestrians, vehicles, animals and objects affected by forces (such as thrown or kicked items).  For now, the research focus will be on the detection of behaviours exhibited by basic 3D shapes acting as agents $R = \{ r_n \}^N_{n = 1}$, where $N$ is the total number of agents within the scene.

An anomalously-acting agent exhibits behaviour that is a deviation from the expected behaviours of other agents in a scene.  In other words, the behaviour shown by an anomalous agent in a crowd is individualistic; the agent has not adopted the behaviour of the crowd \cite{crowd_detection}.

The challenge in anomaly detection comes from the unbounded space of behaviours and physical attributes displayed by an agent that can constitute anomalous behaviour \cite{ucf_crime_dvs}.  A simulator of anomalous behaviour seeks to represent the infinite space of anomalous behaviours as a discrete set that capture the most salient features of agents in the scene.  For example, movement, position, size and rotation are key characteristics that will be displayed by all agents in a scene.  Modelling  attributes such as colour are currently less important as event-based cameras are currently limited to monochromatic capture \cite{dvs_monochrome}.  Therefore, this feature of the behaviour space could be omitted from a simulated anomaly model.

Let $B_g = \{ b_{g,j} \}^\infty_{j=1}$ be the unbound space of behaviours and physical features that can be displayed by an agent of category $g \in G \subset \mathbb{R}$, where $G$ is the total number of agent classes.  The class of an agent is associated here with a different 3D object that defines the shape of an agent; cubes, spheres, cones etc.  

We aim to reduce the dimensionality of this set such that the discrete set of behaviours for a class $A_g = \{ a_{g,j} \}^J_{j=1} \subset B_g$, where $J \in \mathbb{Z}$ is the number of behaviours to be modelled.  An agent is parameterised by its shape and the bound set of behaviours associated with that class; $r_n \{ g; a_g \}$.

Individual behaviours for agents can be modelled using normal distributions.  Central limit theorem states that a large group of independent samples tends toward a normal distribution, regardless of the original distribution of values \cite{clt}.  Therefore, in a highly-populated scene of agents exhibiting a common behaviour of different magnitudes, such as movement along an axis at variable speeds, the distribution of this behaviour will tend towards a normal curve with a mean value $\mu$ and standard deviation value $\sigma$.  Behaviours $a_{g, j}$ can then be formally defined as Eq. \ref{eq:behaviours}, where $m = \mathcal{N}(0, 1)$ is a sample from a zero-mean unit-variance normal distribution.

\begin{equation}
    \label{eq:behaviours}
    a_{g,j} = \mu + \sigma m
\end{equation}

The normalised likelihood (P-value) $\bar{r}_n \in \mathbb{R}^{(0, 1)}$ of an agent existing with its set of behaviours can be found by evaluating the cumulative distribution function $P$ on each of its members, then transforming the sum of these values by the chi-squared distribution function $\hat{P}$.

\begin{equation}
    \label{eq:discrete_behaviours}
    \begin{split}
        \bar{r}_n &= \hat{P} \left( \sum^J_{j=1} -2c \cdot \log \left( P(a_{g, j}) \right) \right), \\
        c &= 1 \text{ if } \sigma > 0 \text{ otherwise } 0 
    \end{split}
\end{equation}

The probability of each agent class $g$ appearing in a scene can also be parameterised as a set of normalised values $W = \{ w \in \mathbb{R}^{(0, 1)} \}^G_{g=1}$.  The relative likelihood $\bar{w}_g$ of an agent existing is given in Eq. \ref{eq:agent_likelihood}.

\begin{equation}
    \label{eq:agent_likelihood}
    \bar{w}_g = \frac{w_g}{\sum^G_{h = 1} w_h}
\end{equation}

The overall likelihood $\omega_n$ for an agent of class $g$ existing with its behaviour values can then be as in Eq. \ref{eq:overall_likelihood}.

\begin{equation}
    \label{eq:overall_likelihood}
    \omega_n = \bar{w}_g \bar{r}_n 
\end{equation}

An anomaly can be defined by comparing $\omega_n$ to a normalised threshold value $v$; $\omega_n \leq v \in \mathbb{R}^{[0, 1]}$.  

An important limitation of this model is the assumption that events are non-causal and independent, which is not accurate to real-world situations.  Given the abstract nature of the simulations involved, we have deemed this limitation acceptable within the current scope of ANTShapes.

\subsection{Crowd Behaviour}
\label{ss:crowd}

The normal sampling method can be adapted as in Eq. \ref{eq:fuzziness} so crowd-like behavior emerges in the scene.  This function shifts normally-sampled values $\mathcal{N}$ to center around $v$.  It then applies a symmetrical curve to enhance the appearance of anomalies against normal-acting agents.  An absolute cubic curve is used as the parabola graphed by this function is steep, which separates anomalous and normal-acting agents effectively.  Anomalous behaviours then appear exaggerated against normal behaviours.  

\begin{equation}
    \label{eq:fuzziness}
    \begin{split}
        \widetilde{\mathcal{N}} &= 
        \begin{cases} 
            (\mathcal{N} + v)^3 & \text{if } \mathcal{N} \leq -v \\
            (\mathcal{N} - v)^3 & \text{if } \mathcal{N} \geq v \\
            0 & \text{if } -v < \mathcal{N} < v
        \end{cases}
    \end{split}
\end{equation}

A \say{fuzziness} coefficient $z$ can be introduced to scale between $\mathcal{N}$ and $\widetilde{\mathcal{N}}$ as in Eq. \ref{eq:fuzziness_scale}.  Increasing the value of $z$ scales towards $\mathcal{N}$, smoothing the bounds between anomalous and normal behaviours.

\begin{equation}
    \label{eq:fuzziness_scale}
    \widetilde{m} = z\mathcal{N} + (1-z)\widetilde{\mathcal{N}}
\end{equation}

This value $\widetilde{m}$ is substituted into Eq. \ref{eq:behaviours} when sampling values defining the visual feedback of behaviour assigned to agents in the scene.  Eq \ref{eq:behaviours} uses samples $m$ drawn from $\mathcal{N}$ rather than transformed values $\widetilde{m} \in \widetilde{\mathcal{N}}$ to ensure previous probabilistic estimations of agent existence likelihood are accurate.

\section{Dataset Generation with ANTShapes}
\label{ss:dataset}

Within ANTShapes, the definitions of simulated anomalies can be tightly controlled by parameterisation of the mean $\mu$ and standard deviation $\sigma$ values for each normal distribution that define agent behaviours.

\begin{figure}[h!]
    \centering
    \includegraphics[width=\columnwidth]{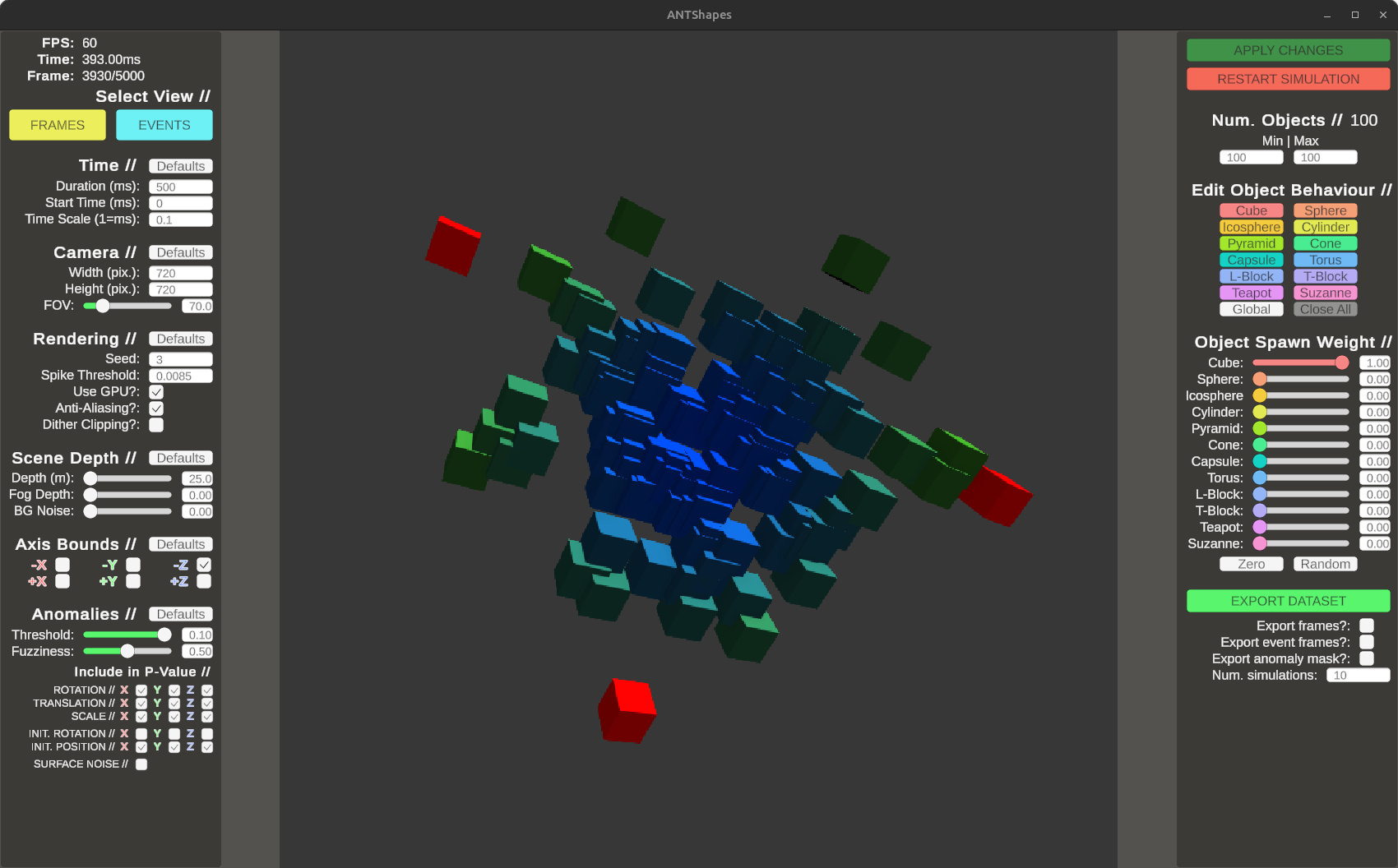}
    \caption{A screenshot of ANTShapes.  The colour of an object indicates its P-value and the likelihood of its existence in the scene by extension.  Blue objects occur relatively frequently, with greener objects being increasingly rarer.  Agents that cross the threshold into anomalies are coloured red.  Here, anomalies are defined based on position.}
    \label{fig:antshapes}
\end{figure}

\subsection{Object Properties}

The following object behaviours are parameterised by normal distributions in ANTShapes:\\

\begin{itemize}
    \item Translation*
    \item Rotation*
    \item Scale*
    \item Initial rotation*
    \item Initial position*
    \item Surface texture noise
\end{itemize}

\textit{* = individual distributions for X, Y, Z axes individually.}\\

Each distribution for these behaviours can be assigned individual values of $\mu$ and $\sigma$.  Samples are taken independently from each distribution upon an agent's creation.  Each behaviour where $\sigma > 0$ can be included or excluded from anomaly definitions according to the needs of the user.  Alternatively, behaviours can be set to constant values of $\mu$ by assigning $\sigma = 0$.

For example, objects can be allowed to rotate and translate around the X, Y and Z axes at independent speeds for each axis.  Here, there are six degrees of freedom which contribute towards anomalous behaviour definitions; one distribution per axis for the rotation and translation behaviours.  Any of these six distributions can be optionally ignored from the agent's P-value calculation (Eq. \ref{eq:discrete_behaviours}) to focus anomaly definition towards desired attributes.  If the user would only like translation speed exceeding a threshold to be defined as anomalous behaviour in the above example, rotation can be ignored to reduce the degree of freedom in the anomalous behaviour space from six to three, simplifying the definition of anomalies in this example.  This allows simulated agents to exhibit multiple complex behaviours for richness and variety in datasets produced by ANTShapes, with freedom for the user to target specific agent properties for anomaly labelling.

ANTShapes also features 12 distinct object classes:\\

\begin{itemize}
    \item Simple flat-faced geometries (cuboids, isospheroids, pyramids, L-block and T-block \say{Tetris-style} shapes)
    \item Simple smooth-faced geometries (spheroids, cones, capsules, cylinders, toruses)
    \item Complex geometries (Utah teapot and Blender's Suzanne monkey head graphical reference models \cite{utah, blender})\\
\end{itemize}

Behaviours can be assigned globally, acting on all agents, and to individual agent classes defined by their shapes.  The inclusion of Gaussian surface noise as an agent behaviour property can reduce the close-up appearance of curved faces as flat-faced meshes.  This effect, a by-product of representing a continuous smooth surface as a discrete surface, is mitigated by applying normal-mapping on the object surface.  Edges between faces are obscured as simulated light is scattered in random directions by the noise texture, rather than purely according to the normals of each face.

Additional rendering options such as the amount of Gaussian background noise, camera field-of-view, depth fog and number of objects in a scene can be definied by the user.  Objects leaving the scene by moving out of the camera frame are replaced by new objects that move into frame according to their translation vectors.  This can be optionally disabled such that agents bounce away from the edges of the camera viewport, forcing objects to remain in the scene.

The user can also change the temporal scale of the simulation.  By default, there is 1ms between simulated frames, which approximates the time-scales seen in frame-based cameras.  Microsecond scales can also be produced to mimic the behaviour of event-based sensors more accurately by lowering the time scale parameter in the software.  Agent behaviours are unaffected by fluctuations in FPS due to rendering differences between machines for consistent simulation.

\subsection{Event Generation}

Spiking events are calculated by comparing the monochromatic pixel intensity at each $x, y$ position of a simulated frame $\mathbf{f}^t$ to the previous frame $\mathbf{f}^{t-1}$ to produce a matrix of difference values $\mathbf{d}_{X \times Y}^t$ as in Eq. \ref{eq:spike_events}.  Bipolar hysteresis gating is applied to each pixel $d_{x, y}^t \in \mathbf{d}_{X \times Y}^t$ to detect changes in intensity compared to a threshold $\pm \beta$.  This produces a matrix of events $\mathbf{e}_{X \times Y}^t$ where increases in brightness are represented by values of 1, decreases by 0 and no meaningful change by 0.5.

\begin{equation}
    \label{eq:spike_events}
    \begin{split}
        \mathbf{d}_{X \times Y}^t &= \mathbf{f}^t - \mathbf{f}^{t-1}, \\
        f(d_{x, y}^t) &=
        \begin{cases} 
            0 & \text{if } d^t_{x, y} < -\beta \\
            1 & \text{if }  d^t_{x, y} > \beta \\
            0.5 & \text{otherwise,}
        \end{cases} \\
        \mathbf{e}_{X \times Y}^t &= \left[ f(d_{x, y}^t) \right]_{x = 1, y = 1}^{X, Y}
    \end{split}
\end{equation}

Visually, the spatiotemporal matrix $\mathbf{e}^{T \times X \times Y}$ can be represented as a $X \times Y$-resolution animation of $T$ frames in length.  Black, white and grey pixels represent values of 0, 1 and 0.5 respectively, as illustrated in Figure \ref{fig:antshapes_spikes}.

\begin{figure}[h!]
    \centering
    \includegraphics[width=\columnwidth]{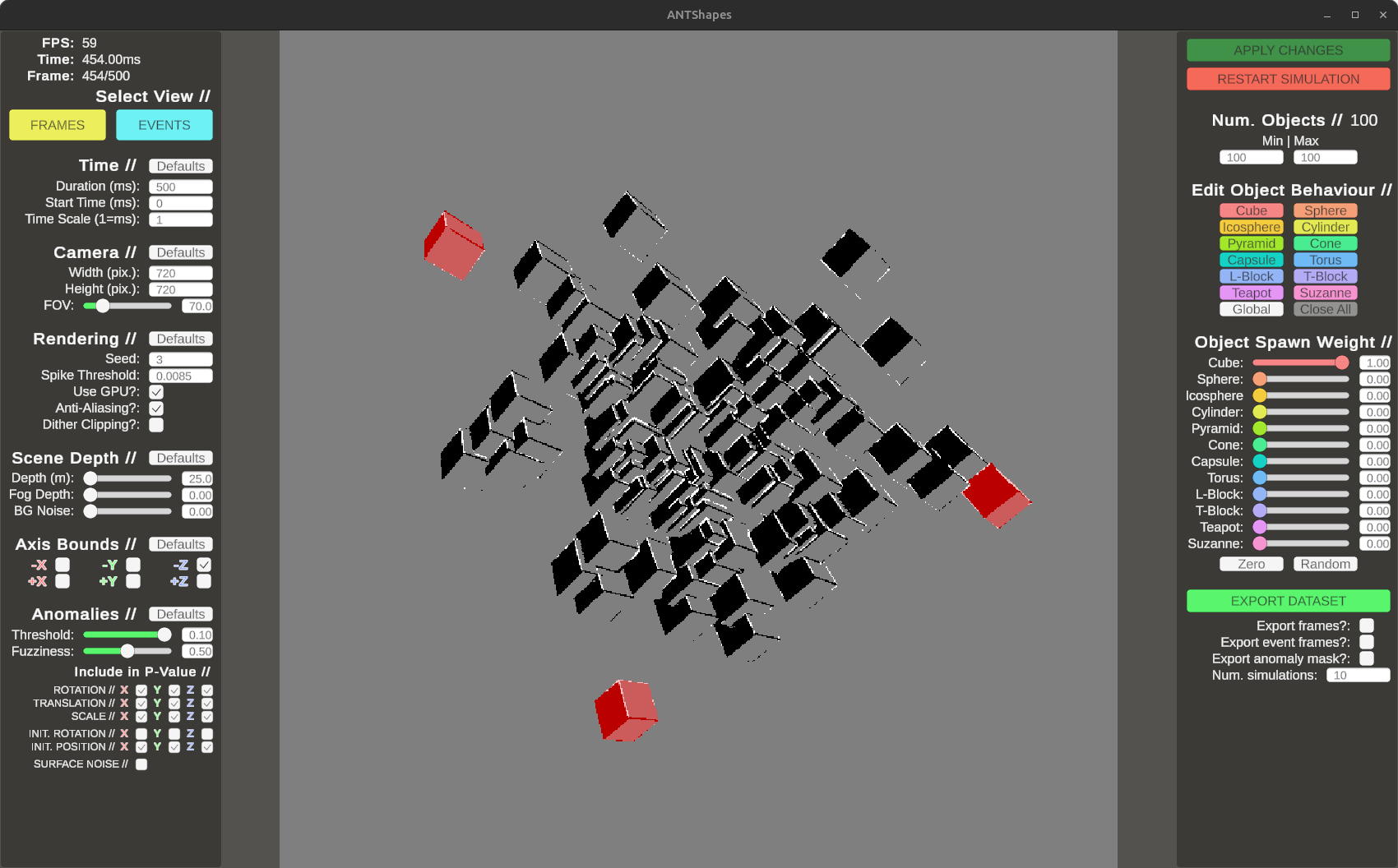}
    \caption{Event representation of the scene in Figure \ref{fig:antshapes}.  Anomalies are defined on the distance of agents from the center of the scene, which are are masked in red.  Each agent is rotating by an equal value on all three axes which produces spiking events.}
    \label{fig:antshapes_spikes}
\end{figure}

As a consequence of ignoring causal behaviour in the anomaly definition model behind ANTShapes, agents are allowed to intersect each other.  It must be noted that future neuromorphic simulators modelling realistic scenes should incorporate causal behaviour into the system for consistency with the real-world inspired approach to dataset generation.  This factor was excluded in ANTShapes as it was deemed out of scope for the simple SNN evaluation purposes that the tool was developed for.

\section{Summary \& Further Work}

Although a comprehensive dataset simulator has been created, ANTShapes is still in the early stages of development at the time of writing.  Therefore, synthesized datasets have not been tested extensively using SNN architectures and the potential of ANTShapes as a dataset simulator is yet to be properly established in practice.  It is the priority of the authors to understand this potential before core development on the software progresses.

As noted in Section \ref{ss:dataset}, the model proposed for anomaly modelling in ANTShapes is a highly simplified model of real-world anomalies that assumes all activity in the scene is independent.  There is interest in extending the anomaly definition model in ANTShapes to more realistic scenes populated by simulated human agents.  For example, non-abstract procedural views of streets could be generated in Unity and populated by human-shaped agents and causality can be modelled by simulating the flow of a crowd and monitoring agent adherance to this flow.  Disruptions in flow will affect the overall state of the system, producing a model of a real-world crowd in an urban environment.

One potential way ANTShapes can be extended is by simulating stereoscopic vision.  Currently, by simulating depth fog and background noise in ANTShapes, we approximate the loss of visual information according to distance.  However, true modelling of binocular vision would allow for a more accurate representation of agent depth to be encoded in simulated datasets.  More generally, camera position and ego-motion could be developed as camera position is currently static and cannot be modified.  Studies conducted using SNNs trained on ANTShapes datasets will guide further development of the software as strengths and weaknesses of the simulator are identified.

\begin{figure}[h!]
    \centering
    \includegraphics[width=0.49\columnwidth]{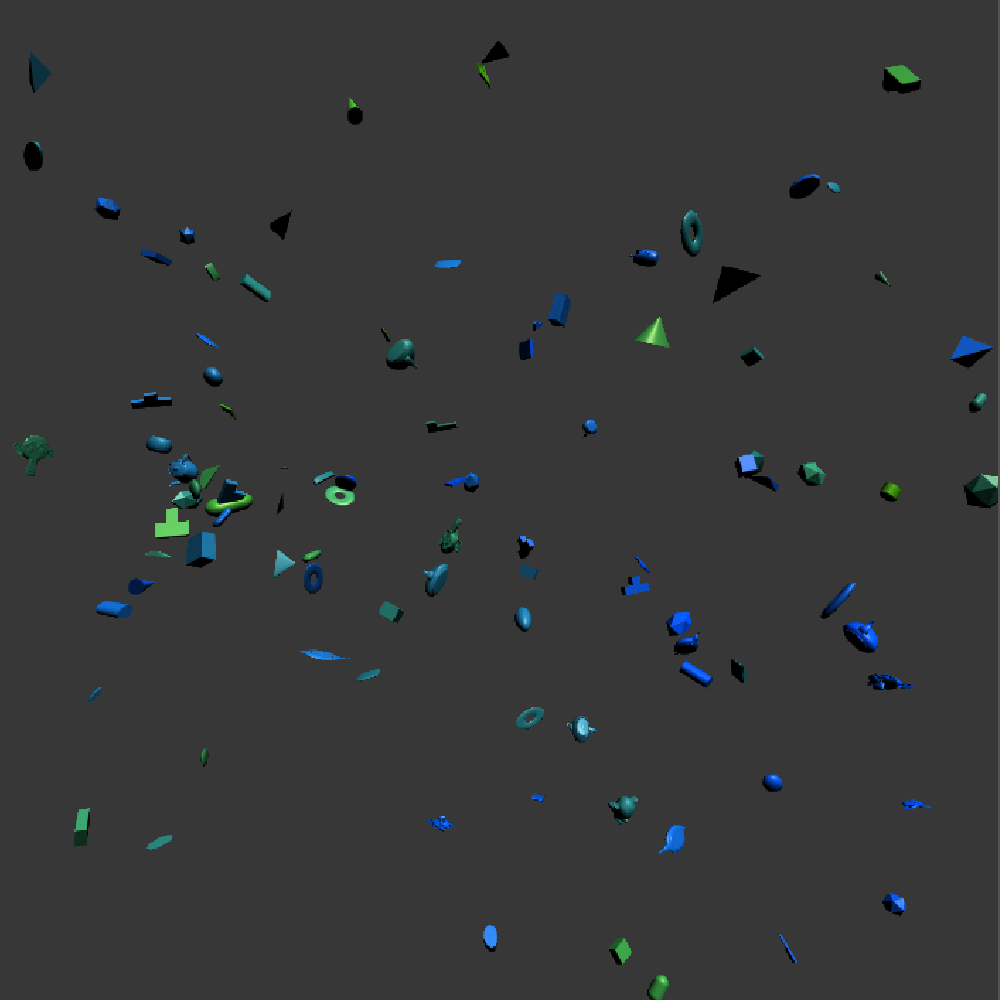}
    \includegraphics[width=0.49\columnwidth]{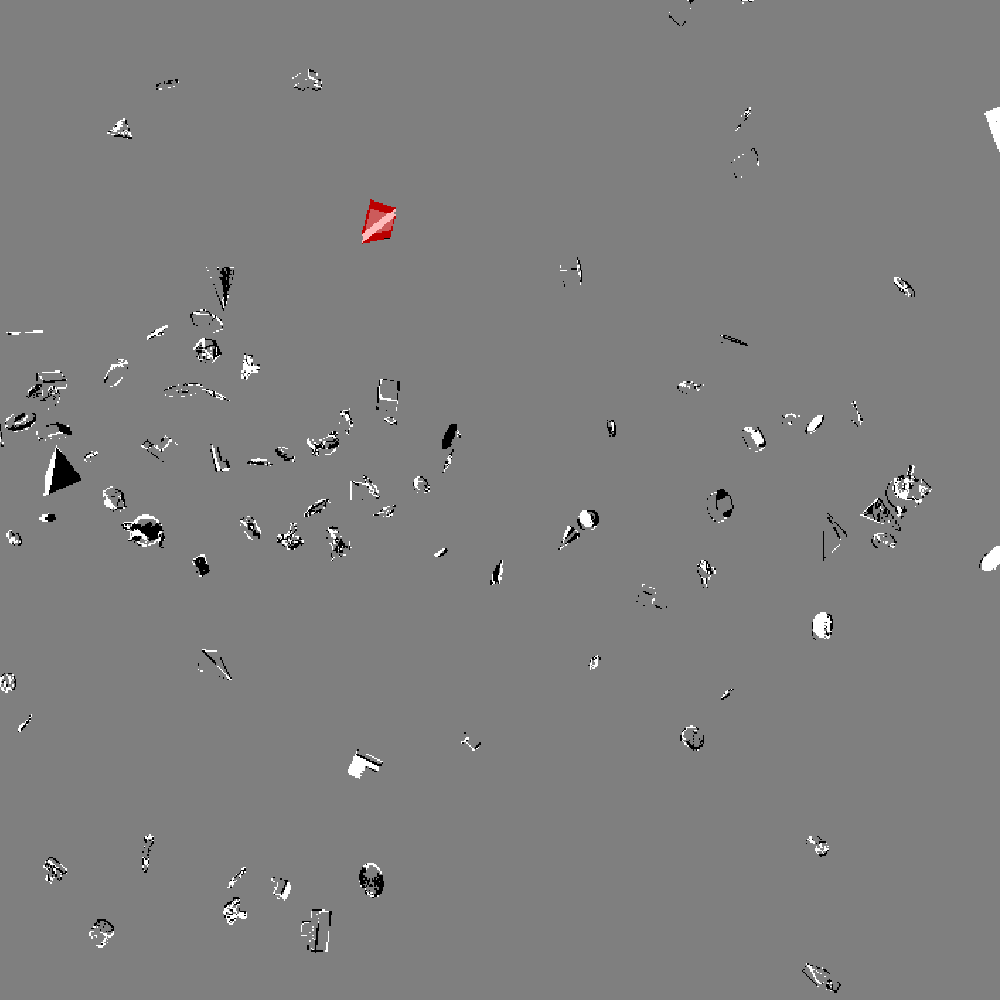}
    \caption{Frame- and event-views from two different temporal moments within a complex scene populated by many shapes with randomised behaviours.  Anomalies were defined as fast-moving objects to produce this example.}
    \label{fig:antshapes_spikes}
\end{figure}

\section{Data Availability}

ANTShapes can be accessed from its GitHub repository: \texttt{\url{https://github.com/EDGYOrganism/ANTShapes}}.  Builds for Linux and Windows are available  with the Unity 6.1 project source.

\printbibliography

\end{document}